%% file: main.tex
\documentclass{article}

\usepackage[english]{babel}

\usepackage[letterpaper,top=2cm,bottom=2cm,left=3cm,right=3cm,marginparwidth=1.75cm]{geometry}

\usepackage{amsmath}
\usepackage{graphicx}
\usepackage[colorlinks=true, allcolors=blue]{hyperref}

\usepackage{graphicx}

\usepackage{tikz}
\usepackage{comment}
\usepackage{amsmath,amssymb} 
\usepackage{color}

\usepackage{hyperref}
\usepackage{bm}
\usepackage{xcolor}
\usepackage[linesnumbered,ruled,vlined]{algorithm2e}
\usepackage{times}
\usepackage{epsfig}
\usepackage{graphicx}
\usepackage{multirow}
\usepackage{verbatim}
\usepackage{enumitem}

\title{Maximum Entropy Baseline for Integrated Gradients}
\author{Hanxiao Tan \\
AI Group, TU Dortmund, Germany \\
hanxiao.tan@tu-dortmund.de}
\date{}

\DeclareMathOperator*{\argmax}{argmax}

\newcommand{\approptoinn}[2]{\mathrel{\vcenter{
  \offinterlineskip\halign{\hfil$##$\cr
    #1\propto\cr\noalign{\kern2pt}#1\sim\cr\noalign{\kern-2pt}}}}}

\newcommand{\appropto}{\mathpalette\approptoinn\relax}
    
\newcommand{\beginsupplement}{%
        \setcounter{section}{0}
        \renewcommand{\thesection}{S\arabic{section}}%
        \setcounter{table}{0}
        \renewcommand{\thetable}{S\arabic{table}}%
        \setcounter{figure}{0}
        \renewcommand{\thefigure}{S\arabic{figure}}%
        \setcounter{equation}{0}
        \renewcommand{\theequation}{S\arabic{equation}}%
     }

\begin{document}

\maketitle

\begin{abstract}
Integrated Gradients (IG), one of the most popular explainability methods available, still remains ambiguous in the selection of baseline, which may seriously impair the credibility of the explanations. This study proposes a new uniform baseline, i.e., the Maximum Entropy Baseline, which is consistent with the "uninformative" property of baselines defined in IG. In addition, we propose an improved ablating evaluation approach incorporating the new baseline, where the information conservativeness is maintained. We explain the linear transformation invariance of IG baselines from an information perspective. Finally, we assess the reliability of the explanations generated by different explainability methods and different IG baselines through extensive evaluation experiments.
\end{abstract}

\input{1_Introduction}
\input{2_Relatedwork}
\input{3_MaxEntropyBaseline}
\input{4_EntropyAblationEvaluation}

\input{5_Evaluations}
\input{6_Conclusion}

\bibliographystyle{alpha}
\bibliography{egbib}

\clearpage
\input{7_Supplementary}
\end{document}

%% file: 1_Introduction.tex
\section{Introduction}

Integrated Gradients (IG) \cite{pmlr-v70-sundararajan17a} is one of the most widely used explainability methods at present, which illustrates the attribution of each pixel in the input $x$ to the corresponding prediction. IG is a gradient-based approach that addresses the gradient saturation problem \cite{sundararajan2016gradients} of vanilla gradients by integrating them from a chosen baseline $x'$. IG is formulated as:

\begin{equation} \label{IGformular}
    IG_i=(x_i-x_i')\cdot \int_{\alpha=0}^{1}\frac{\partial F(x'+\alpha (x-x'))}{\partial x} d\alpha 
\end{equation}

The baseline is one of the parameters of IG, which strongly impacts the performance of the generated explanations. The existing studies, although specifically suggesting different baselines, are almost unanimous in their overall definition: missingness, i.e., the input that disables the model from capturing information. Currently the most prevalent baselines are: 1) \textit{Zeros}: filling all input units with zeros. 2) \textit{Black (white) vectors}: replace all units with the minimum (maximum) of the current input. 3) \textit{Random initializations}. In addition, \cite{sturmfels2020visualizing} complements several possible alternatives: 4) \textit{Max-distance pixels (Xdist)}: their study reveals that the explanations generated by IG tends to be more sensitive to pixels that differ significantly from the baseline, while ignoring those that are similar, even if they are located inside the object to be explained. Thus, they propose Xdist, i.e., replacing each pixel with the point that is farthest apart in chromaticity space. 5) \textit{Blurring}: information is eliminated by blurring the input with kernel functions. 6) \textit{Random noises}: as a supplement of 3), random noise of different distributions (e.g. Gaussian and uniform) can be substituted or added to the input. 7) \textit{Average over datasets}: random samples are drawn from the training dataset and the average is taken. Notably, while 1)-4) assume that the features are independent of each other, 5)-7) incorporate correlation between features. However, we argue that the existing concept "missingness" lacks quantitative metrics, which leads to the ambiguity in the choice of baselines.

Another cause of difficulty in baseline selection is the lack of ground truth, such that evaluating the performance of explanations is challenging. Ablation test \cite{sturmfels2020visualizing} is a simple and comprehensible solution. The core principle is that if the points with positive attributions are ablated, the model drops its confidence in the prediction and vice versa. Nevertheless, we believe that the existing ablation tests suffer from insufficient reliability, primarily due to 1) No uniform replacements as substitutes for the original pixels. For instance, both \cite{bach2015pixel} and \cite{samek2016evaluating} employ ablation tests to evaluate performances of the explanations. However, with respect to the substitution pixels, \cite{bach2015pixel} chooses the opposite number of the current pixel while \cite{samek2016evaluating} samples random numbers from uniform distribution. This raises the question of which substitute is more appropriate, or if there is a better alternative. 2) No guarantee of "Neutral": after flipping selected pixels, there is no metric for whether information residues survive. Both of the above points may significantly impact the reliability of explanation assessment.

To address the aforementioned weaknesses, this work re-examines the role of the baseline from the perspective of information and refines the ablation test to satisfy conservativeness. We observe and propose our conjecture on a synthetic toy dataset and validate it on MNIST handwriting dataset. Moreover, we also compare ours with other existing baselines on CIFAR10. Our contribution is primarily summarized as follows:

\begin{itemize}
    \item We propose a new baseline that remains "uninformative" by definition, and by constructing toy datasets with transparent ground truth explanations, we observe that our baselines closely approximates the true value.
    
    \item We propose an improved ablation test for evaluating explanations that specifies a uniform substitution of ablated pixels and simultaneously satisfies the conservativeness of the information quantity.
    
    \item We explain the invariance of linear transformations of the IG (proposed by \cite{kindermans2019reliability}) from the perspective of the entropy curves.
    
    \item We validate our baseline on different datasets, and compare popular saliency map-based explainability methods with proposed evaluation metric.
\end{itemize}

The paper is structured as follows: Section \ref{related_work} introduces the current studies on IG baselines and ablation tests. Sections \ref{Methods: Xentr baseline} and \ref{Methods: Xentr ablation} elaborate the experiments related to the proposed baseline and ablation test, respectively. Section \ref{quantitative_evaluation} presents the results of the experimental evaluation of the proposed methods.

%% file: 2_Relatedwork.tex
\section{Related Work} \label{related_work}

\textbf{Explainability methods}. 
Existing explainability approaches can be generally categorized into two classes, i.e., surrogate model-based black-box methods and gradient-based white-box methods. Surrogate model-based approaches typically simulate the performance of the original model with a simple, explainable one, thereby achieving explainability, such as LIME \cite{ribeiro2016should}, Anchor \cite{ribeiro2018anchors} and KernelShap \cite{lundberg2017unified}, etc. Due to the simplified structure, the performance of the agent model can hardly be identical to the original one, which leads to that the majority of those methods approximate only in the vicinity of the instance to be explained (locality). Moreover, perturbation is challenging: the plausibility of training a surrogate model with unreasonable perturbation examples is questionable \cite{molnar2020interpretable}. Another series is the gradient-based approaches that require accessing the internal construction of the model. As a pioneer of the saliency map, \cite{simonyan2014deep} (Vanilla gradients) derives the prediction attribution by observing the gradient changes of individual pixels. Several variants have been proposed aiming at improving the methodology, e.g., Simple Taylor Decomposition (input $\times$ gradients) \cite{montavon2017explaining}, Guided Back-propagation \cite{springenberg2014striving} and SmoothGrad \cite{smilkov2017smoothgrad}, etc. IG \cite{pmlr-v70-sundararajan17a} is in one of the subseries, namely the path method. Vanilla gradient may be saturated at the current input and the derived significance map may not reflect the real attributions \cite{sundararajan2016gradients}. IG addresses this issue by integrating the gradients from an "uninformative" initialization to the current input. Therefore, the definition and selection of this start point is relevant to the plausibility of its explanations.

\textbf{Investigation of IG baseline}. So far, there are no sufficient researches for IG baselines. Several studies have pointed out that the choice of baseline impacts dramatically on the explanation performance \cite{qi2019visualizing,sturmfels2020visualizing,kindermans2019reliability}, with the appropriate baseline generating clearer attribution maps and the opposite distorting them. \cite{kindermans2019reliability} also indicates that a number of specific baselines (e.g., zeros) do not satisfy linear transformation invariance and are hence unreliable. The proposer of IG provides a definition of baselines \cite{pmlr-v70-sundararajan17a}, i.e., when the model prediction is neutral, and suggest that for most networks the zero baseline is applicable, while also specifying different applications for networks in different domains, e.g., the black background for image networks and the zero embedding vector for NLP tasks. As a complement, \cite{qi2019visualizing} proposes several additional baselines that are intuitively feasible, such as the Maximum Distance and Blurred baselines. Although these baselines are either consistent with human intuition or address certain deficiencies of existing ones, there is no convincing argument that the proposed baselines are "uninformative".

\textbf{Ablation test}. In LRP \cite{bach2015pixel}, the ablation test is first proposed as a validation for its explainability method, which has being followed or expanded in several studies relating to feature-wise attributions \cite{samek2016evaluating,montavon2019gradient,zheng2019pointcloud,tan2022surrogate}. However, \cite{hooker2019benchmark} pointed out that the method suffers from the potential problem that ablating a single feature might impair the feature correlation and data distribution, and they suggest that a new model should be retrained after ablating for accuracy validation (Remove \& Retrain). Again, this also leads to debates: should the explanation be faithful to the data or the model \cite{sundararajan2020many,janzing2020feature}? Leaving aside the controversy about feature correlation, we expose another potential risk of elimination experiments from the information perspective, i.e., non-conservation, and propose a corresponding solution. 

%% file: 3_MaxEntropyBaseline.tex

\section{Max entropy baseline} \label{Methods: Xentr baseline}

The ideal baseline is the input that "is neutral" \cite{pmlr-v70-sundararajan17a} and "contains no information" \cite{sturmfels2020visualizing}. However, measuring the amount of information embedded in the inputs is challenging since most models are black boxes and only final predictions are observable. On the other hand, we argue that the referent of "neutrality" is ambiguous. Existing referents for the baseline are $3$ categories: uninformative for humans, data and models. Most baseline choices are for humans, such as zero, black and random baselines: it is intuitively assumed \textbf{by humans} that these values do not yield any information. Nevertheless, for the model to be explained, this may be problematic (e.g., a classifier that distinguishes between black and white images). Subsequently, those baselines for the dataset (e.g., Average of training data) also raise controversy: whether the explanation should be true to the data or the model \cite{sturmfels2020visualizing,sundararajan2020many,janzing2020feature}. So far, there is hardly any baseline for the model being proposed. To address the aforementioned flaws, we introduce the entropy of logits as a metric for quantifying information residual in the model. We denote the proposed baseline as $B_{Xentr}$, and is formulated as: 
\begin{equation} \label{New_baseline}
    B_{Xentr} = \argmax_x H(Softmax(f_l(x))
\end{equation}
where $f_l$ denotes the logits of the model and $H(*)$ denotes the entropy function \cite{shannon1948mathematical}:

\begin{equation}
    H(A) = -\sum_{i=1}^{n} P(a_i)logP(a_i)
\end{equation}

\begin{figure*}
    \begin{centering}
    \includegraphics[width=1\textwidth]{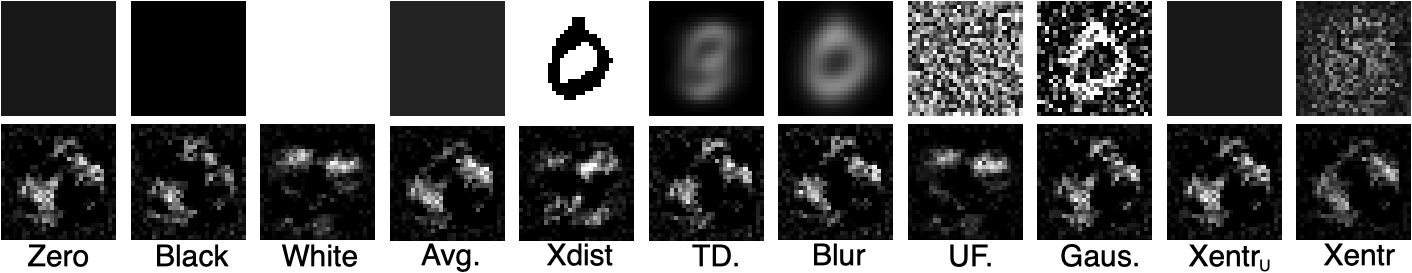}
    \caption{Visualization of different baselines and corresponding explanations. From left to right: Zero, black, white, average of current instance, Max Distance, average of training data, blurred, uniform distributed, Gaussian distributed, uniform and non-uniform Maximum Entropy baselines.}
    \label{visu_diff_baselines}
    \end{centering}
\end{figure*}

Our newly proposed baseline has the following advantages: 
\begin{itemize}
    \item \textit{Input independent}. Several baselines incorporate (or incompletely ablate) information from input instances to attain better visualizations. We argue that this violates the definition of baselines. For example, The Maximum Distance baseline (Xdist) \cite{sturmfels2020visualizing}, which calculates the maximum colorimetric distance of each pixel from the input, moderates the phenomenon of "attribution disappearance" in the explanations. Nevertheless, this baseline obviously includes extensive information about the input instances, such as object outlines (see figure \ref{visu_diff_baselines}).
    
    \item \textit{Feature correlation Incorporated}. The plausibility of generating explanations based on the assumption of feature independence remains questionable. Owing to the strong correlation between features, learning a certain distribution based on the dataset is a more convincing solution \cite{sturmfels2020visualizing}. Alternatively, the baseline we proposed is derived from the optimization of a trained model, which itself possesses the distribution of the dataset.
    
    \item \textit{Linear transformation invariant}. \cite{kindermans2019reliability} suggests that the generated explanation should remain constant when the input and the model undergo a uniform linear transformation (all pixels transformed with the same amplitude). A portion of the baselines fails this test (e.g., zero baseline), while several others (e.g., black baseline and ours) survive. In addition, we elucidate from an information perspective why all baselines fail the non-uniform linear transformation test in section \ref{Method:LT invariance}.
    
    \item \textit{Computational simplicity}. Being input independent, only a simple gradient ascent procedure is required for each model to obtain $B_{Xentr}$, and is applicable to all predictions made by the model.
\end{itemize}

Figure \ref{visu_diff_baselines} shows all currently available IG baselines and their generated explanations. Ahead of assessing the validity of the proposed baseline, we show our observations with respect to the correlation between the entropy of logits and the explanations of IG. In section \ref{Method:toydataset} we show experimental observations in a tabular toy dataset, and in section \ref{Method:MNIST} we extend the findings to the MNIST handwriting dataset. Note that for achievable computational complexity, in this section the baseline is treated as a uniform value ($B_{Xentr_u}$). 
    
\subsection{Warm-up: tabular toy datasets} \label{Method:toydataset}
\begin{figure}
    \begin{centering}
    \includegraphics[width=0.475\textwidth]{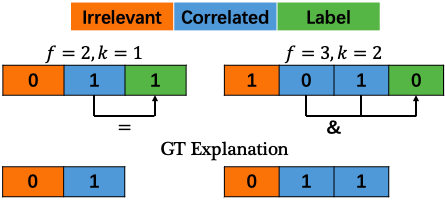}
    \caption{The structure of the tabular toy data set, where $f$ and $k$ denote the total amount of features and the number of those relevant to the labels, respectively. The last row illustrates the ground truth explanations.}
    \label{description_toy}
    \end{centering}
\end{figure}

The lack of ground truth explanations is one of the obstacles to investigating baselines, we thus artificially create tabular toy datasets whose ground truth explanations are available. An overview of the dataset is shown in figure \ref{description_toy}. Our dataset involves $3$ parameters, namely $f$, $k$ and $c$, representing the total number of features, the number of features correlated with the label and the total number of label categories, respectively. For each individual dataset, there is at least one feature correlated with the label ($k\geq1$). Hence, we can obtain a unique ground truth explanation for this dataset, i.e., an attribution of 1 for the features relevant to the labels and 0 for the opposite. Note that when $k > 1$, the labels must be computed jointly based on all relevant features, otherwise the model may learn only a part of them, leaving the ground truth explanations redundant. Each dataset consists of $10^4$ instances (most of the data are duplicates due to the limited combination of features), of which $80\%$ are used as training data and the rest are for testing. A simple two-layer fully connected network is trained, which achieves $100\%$ accuracy on both the training and test sets. We traverse the explanations of all the baselines in the valid data value range with 100-step IG. Subsequently, we measure the gap between the IG-generated and ground truth explanations. As the exact value of the ground truth explanations is not reachable, we quantify the loss with KL-divergence, i.e., features related to the label should possess as large attributions as possible, otherwise should approximate to 0. It is notably that only the results of the instances where all features are 0 as the object being explained are shown, since according to the properties of the dataset, the ground truth explanations should be independent of the input instances, and we practically tested the input instances for all feature combinations and the results are consistent (see figure \ref{sup_4_diff_vectors}). On the other hand, we draw the entropy curve of the model over the valid range of data values. The prediction are fed into a Softmax function to assure that each logits neuron is positive, and then the entropy of this probability vector is calculated.

\begin{figure*}
    \begin{centering}
    \includegraphics[width=1\textwidth]{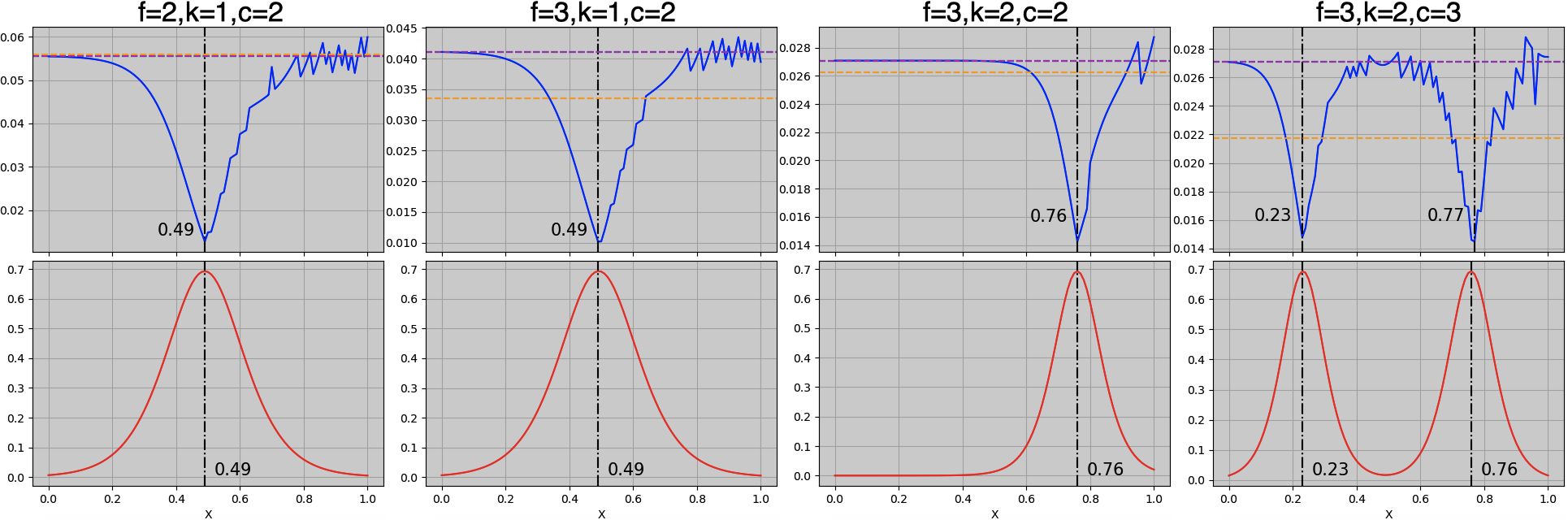}
    \caption{First row: KL-losses of between the IG explanations obtained with the corresponding inputs (x-axis) as baselines and the ground truth explanations. Second row: entropy curves of the model. $f$ denotes the total number of features, $k$ denotes the number of features associated with the label, and $c$ denotes the total number of label categories. We also marked the loss of zero (green), black (cyan), white (purple) and random (orange) baselines (partially obscured by each other, but significantly higher than the loss of $B_{Xentr}$).}
    \label{toydata_res}
    \end{centering}
\end{figure*}

The results are reported in figure \ref{toydata_res}. Interestingly, the loss and entropy curves follow similar trends and reach extreme values at approximate baselines. These results are intuitively correct: the maximum predictive entropy implies the minimum information content of the input vector, which exactly coincides with the characteristics of the baseline in IG.

\subsection{MNIST handwriting dataset} \label{Method:MNIST}

To extend this conclusion to a more general practice scenario, a similar experiment is conducted on the MNIST handwritten dataset. However, the major challenge with real datasets compared to the artificial one is that no prior knowledge is available about which features (pixels) in the instance are decisive for the labels, i.e., the ground truth explanations are not accessible. Existing studies treat those pixels located inside the object as ground truth \cite{simonyan2014deep,zhang2018top}. Nevertheless, no guarantee can be given that the model does not utilize any information from the background when inferring \cite{arras2020ground}. Our alternative is to aggregate multiple explainers to yield hybrid explanations (see section \ref{hybrid_explainer}). The voting system reinforces pixels with identical attribution signs, while the attributions of those ambiguous ones counterbalance each other. Notwithstanding the inability to precisely restore the ground truth explanations, this approximated saliency map captures the trend of the attribution distribution given by the majority of explainers. Before training, all data are transformed to the value domain $[-0.42,2.82]$ for higher accuracy. We train two different types of networks, the FC networks consisting of only fully connected layers and the CNN networks containing convolutional layers. Similarly, we search all the baselines in the valid value range and produce the explanations by a $100$-step IG. Owing to the high dimensionality, KL-loss struggles to reveal the distributional discrepancy, we therefore adopt Spearman's rank correlation coefficient to measure the similarity between explanations. We randomly selected $100$ instances from the test set for each model and record the baselines at minimum loss with histograms. Again, the curve of logits entropy is also plotted.

\begin{figure*}
    \begin{centering}
    \includegraphics[width=1.0\textwidth]{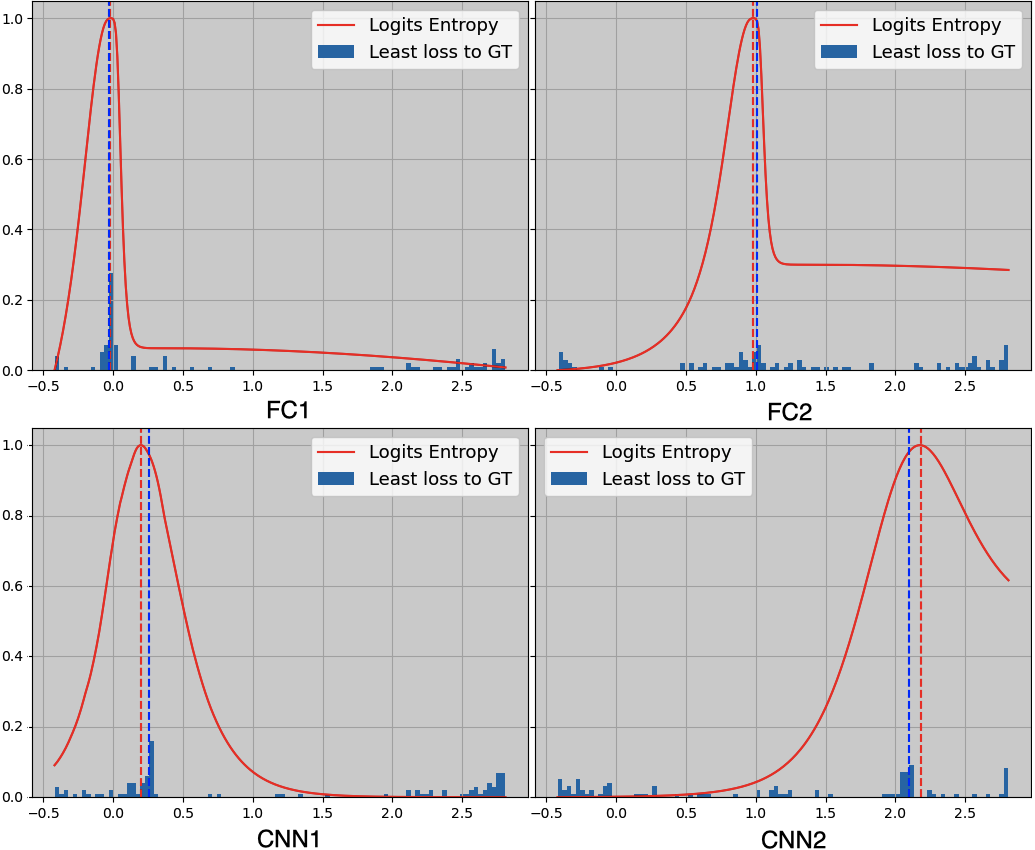}
    \caption{The entropy curves of the logits and density histograms of baselines which achieve the minimum Spearman loss with the (approximated) ground truth explanations. The red and blue dashed lines indicate the maximum values of the logits entropies and density boxes, respectively.}
    \label{4_net_estimation}
    \end{centering}
\end{figure*}

The results are demonstrated in figure \ref{4_net_estimation} and the maximum of both recordings are annotated with a dashed line in the corresponding color. As can be observed from the two plots on the left, the baseline with the lowest loss almost coincides with the input of the maximum logits entropy. To exclude the possibility that the minimum loss baseline is around $0$, two additional models are trained whose input of maximum logits entropy deviate far from the origin point (plots on the right). Although the minimum loss distributions are not perfectly concentrated at the maximum entropy values, significant following offsets can be observed. Additionally, we find that the reason why zero is feasible as a baseline alternative is that the global maximum of the entropy curve for a portion of neural networks lies approximately adjacent to the origin point. Nevertheless, the opposite also exists, for instance, the FC2 and CNN2 in our experiments (figure \ref{4_net_estimation}, top right and bottom right). FC2 is derived from performing an identical linear transformation on the dataset and the bias of the first layer of the network for FC (the experiment is first proposed by \cite{kindermans2019reliability} and is described in detail in the next section), while CNN2 is a convolutional neural network with more sophisticated architectures. The maximum values of their entropy curves deviate from the origin, at which point the zero baseline can no longer be considered as an approximation, while the baseline closest to the approximated ground truth explanations can be observed to be shifting towards the positive direction of the X-axis, rather than remaining at the origin.

\subsection{Linear transformation invariance} \label{Method:LT invariance}
\begin{figure*}
    \begin{centering}
    \includegraphics[width=1\textwidth]{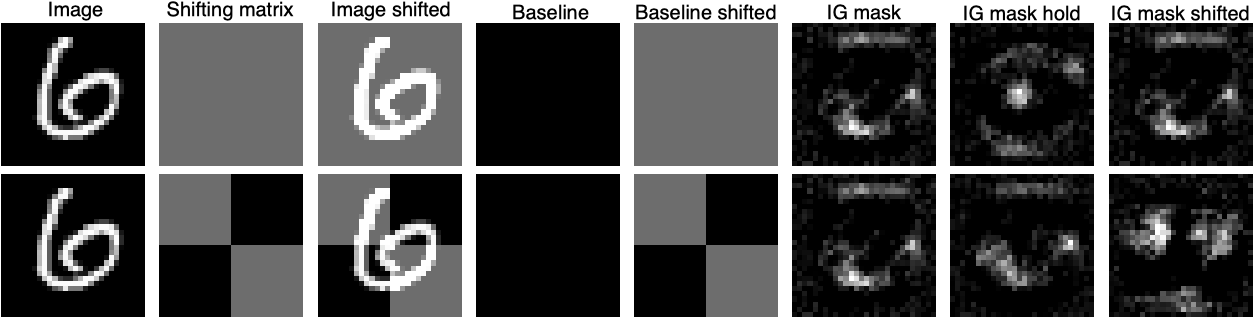}
    \caption{Visualizations of the uniform and non-uniform linear transformations on the instances and baselines. The last three columns are the corresponding explanations. IG mask hold demotes the linear transformation is applied only to the instances, whose baselines remain non-transformed.}
    \label{linear_transform_visu}
    \end{centering}
\end{figure*}

Previous study point out that a fraction of the IG baselines fail the linear transformation test \cite{kindermans2019reliability}. They shift the input instances and the parameters from the first layer of the model with the same linear offset, and subsequently observe whether the explanations generated by the explainer are consistent before and after the transformation. In the experiment, two alternative shifts are adopted as offsets, i.e., the uniform and non-uniform shifts, where the former is equally shifted at each pixel, while the latter is not subject to this restriction. They utilize black and zero baselines for IG, while only the former maintains the same explanation before and after the uniform shift. We reproduce the experiments and present them in figure \ref{linear_transform_visu} (the shift amplitudes, including all uniform shifts and the non-zero pixels of the cross-shaped shifts, are $0.5$). We argue that the black baseline is adaptive and shifts with the input, whereas the zero baseline remains constant, which lacks fairness. Therefore, we transform the zero baseline with the same offset and observe the explanation invariance. The results are shown in the first row where the zero baseline maintains the consistent explanation after the transformation. The second row exhibits the results of the non-uniform shift, and interestingly, when we perform an identical shift for the zero baseline, the explanation is irreducible.

\begin{figure}
    \begin{centering}
    \includegraphics[width=0.6\textwidth]{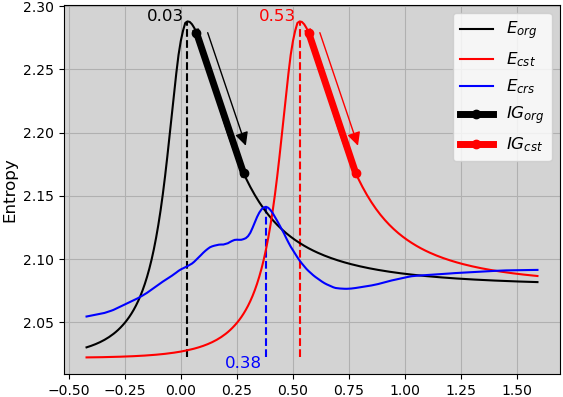}
    \caption{Entropy curves of the model after different shifts, where $org$ stands for no shift, $cst$ stands for a uniform constant shift of the input vector, and $crs$ stands for a cross-shaped shift of the input vector with the midline as the dividing point, $IG_{org}$ and $IG_{cst}$ denote the integral of information from high-entropy initiations to low-entropy destinations (the explanations).}
    \label{reliable_shift}
    \end{centering}
\end{figure}

From the information perspective, the entropy curves may illustrate why the explanations of uniform linear shifts can be reproduced, but not non-uniform ones. Figure \ref{reliable_shift} plots the entropy curves of the three models in a fraction of the data value range. Suppose the entropy curve of the original model is $H(x)$ (black curve), which turns into $H_U(x)$ after the uniform linear transformation (red curve), and the transformation amplitude is $A_s$. The correlation can be easily derived from the diagram:
\begin{equation} \label{infor_qual}
    H_U(x)=H(x-A_s)
\end{equation}

Equation \ref{infor_qual} indicates that the information content of the two models is identical except for a phase difference of $A_s$. However, the model with cross-shaped transformation possesses a severely distorted entropy curve (blue curve). When the original model is explained with IG, the gradient is accumulated from an "uninformative" initiation to the destination to be explained, and an identical path can be found in the uniformly transformed model, with the starting and ending points translated by $A_s$ (e.g., bolded black and red segments), while the integrals are equivalent. In contrast, such a segment is absent in the curve of the model with the cross-shaped transformation. Consequently, models that undergo non-uniform shifts can be treated as entirely novel ones that cannot reproduce the explanations of the original model by simple linear transformations.

%% file: 4_EntropyAblationEvaluation.tex
\section{Ablation-based evaluation methods} \label{Methods: Xentr ablation}

In this section, we elaborate on the flaw of the classical ablation-based evaluating metric for explainability methods, which monitors the prediction activations in \ref{tradition_ablation_test}, and propose a novel one, which targets the entropy as the surveillance \ref{new_ablation_method}.

\subsection{Existing ablation test} \label{tradition_ablation_test}

The ablation approach and its variants are the most commonly used methods to evaluate explanations, which is comprehensively summarized in \cite{sturmfels2020visualizing}. Let $x \in X$ denote an instance containing $n$ features $x_{o}=(x_{o}^1,x_{o}^2,...,x_{o}^n)$ ($X$ is the set of all valid instances), the information quantity it contains before the ablation test is initiated as $I(x)$. After all $n$ features are ablated, the input now is noted as $x_{\phi}=(\phi^1,\phi^2,...,\phi^n)$, where $\phi_i$ denotes the ablated substitutions, and the information quantity it holds is marked as $I(x_{\phi})$, the ablating evaluation can be formulated as:

\begin{equation}
    S_{e} \propto I(X_o)-I(X_{\phi}^p)
\end{equation}
where $S_{e}$ indicates the score of the explainer and $X_{\phi}^p$ indicates the removal of p positively attributed features.

Several studies \cite{bach2015pixel,samek2016evaluating,montavon2019gradient} employed ablation approaches to validate the reliability of explainability methods, while differing slightly in the details. The major distinctions are:

\begin{itemize}
    \item \textit{Surveillance target}. Let $l$, $f$ represent the prediction label and the model to be explained. Frequently monitored objects are collectively defined as "prediction scores" and are divided into 1. the corresponding activations in logits: $f_l(x)$. 2. the corresponding cells in logits after taking Softmax, which can be considered as the probability of the class: $\sigma(f_l(x))$. In these works, they consider the reduction of prediction scores as a result of "information removal" \cite{samek2016evaluating}, the surveillance object is regarded as a measurement of "information quantity", i.e., $I\approx \Delta f_l(x)$ or $\Delta \sigma(f_l(x))$.
    
    \item \textit{Ablated destination}. The substitution value for ablation is also controversial. There are several competitors: zero, minimum value of the instance, reversing the sign of the current pixel \cite{bach2015pixel}, blurring via Gaussian kernel \cite{sturmfels2020visualizing} etc. The purpose of ablation is to conceal information about a specific pixel (or feature), the flipping destinations themselves should therefore carry no information, i.e., $I(x_{\phi}) \approx 0$.
\end{itemize}

However, we argue that such metrics suffer from non-conservation. According to the definition of the ablation test, the information quantity of the ablated input should satisfy:
\begin{equation} \label{infor_const}
    \forall x\in X, I(x)\geqslant I(x_{\phi})
\end{equation}

Note that practically the information quantity is not directly available and the existing methods record the value of logits $f_l$ (or the softmax of the logits $\sigma(f_l)$) as substitutes. Thus, Equation \ref{infor_const} can be rewritten as:
\begin{align} \label{logits_const}
    \forall x\in X,  f_l(x)\geqslant f_l(x_{\phi})\ or \  \sigma(f_l(x))\geqslant \sigma(f_l(x_{\phi}))
\end{align}

\begin{figure*}
    \begin{centering}
    \includegraphics[width=1\textwidth]{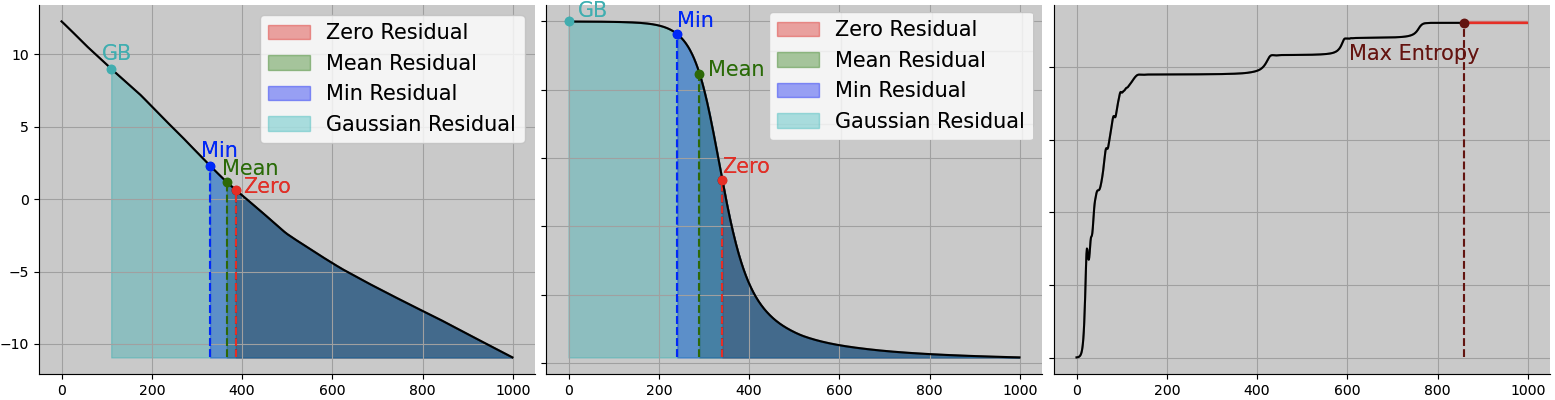}
    \caption{The ablation experiments with the raw logits activations (left), softmax of logits activations (middle) and the entropy of logits as the surveillance target (right), respectively. The x-axis denotes the number of optimization steps and the y-axis represents the value of the corresponding monitored target. The shaded areas with colors represent "information residues".}
    \label{ablation_diff}
    \end{centering}
\end{figure*}

Taking the CNN in Section \ref{Method:MNIST} as an example, we employ a $1000$-step gradient descent algorithm to minimize the value of the surveillance target and record the changing curve. Subsequently, we feed various $x_{\phi}$ into the model, derive the individual predicted values, and annotate them on the curve. Notably, the optimization process is performed in the valid range of the original dataset. According to figure \ref{ablation_diff}, equation \ref{logits_const} is violated by the optimization results, and numerous inputs with lower monitoring target values can be identified in the valid range. If $f_l(x)$ is considered as the information quantity, $x_{\phi}$ suffers from extensive information redundancy (left plot). Compared to the former, $f_l(x_{\phi})$ mitigates this deficiency, while the information residual in $x_{\phi}$ is still visible (middle plot). Experiments show that the ablating test designed on the basis of inputs that are commonly understood by humans as "uninformative" may still contain information and may be problematic.

\subsection{Entropy-based ablation test} \label{new_ablation_method}

To fulfill the conservativeness in Equation \ref{infor_const}, we extend the definition in equation \ref{New_baseline}, that monitors the entropy of logits as the quantitative indicator of information, i.e.

\begin{equation} \label{information_prop}
    I(x) \appropto \frac{1}{H(\sigma(l))}
\end{equation}

The increment of information diminishes the uncertainty of events, which can be expressed by the entropy. The introduction of entropy facilitates the understanding of "information quantity". On the other hand, by redefining the unique ablated destination as $x_{\phi}=B_{Xentr}$, the conservativeness of the ablation assessment is assured

\begin{align}
    I(B_{Xentr}) \approx \frac{1}{H(\sigma(f_l(B_{Xentr})))}
\end{align}

Recalling equation \ref{New_baseline}, which yields
\begin{equation}
    \forall x\in X, H(\sigma(f_l(x))) \leq H(\sigma(f_l(B_{Xentr})))
\end{equation}

and refers to equation \ref{information_prop}, equation \ref{infor_const} holds. To experimentally verify the conservativeness, we depict the entropy curve and mark out $B_{Xentr}$, as shown in Figure \ref{ablation_diff} (right plot). Note that in this plot, the area above the entropy curve represents the information residual, which does not exist since $B_{Xentr}$ is the input that maximizes the entropy.

%% file: 5_Evaluations.tex
\section{Quantitative evaluations} \label{quantitative_evaluation}
In this section, we evaluate the performance of various IG baselines with the proposed ablation test in section \ref{evaluate_baselines}, and assess different explainability methods in section \ref{evaluate_exp_meds}. We conduct evaluation experiments on two different datasets, MNIST and CIFAR10. For MNIST, we train two different models, i.e., fully connected network and convolutional neural network, which achieve $98.2\%$ and $98.5\%$ accuracies on the test set, respectively. For CIFAR10, we train a ResNet18 \cite{he2016deep} network, whose test accuracy is $95.6\%$. During evaluation, for MNIST we evaluate all $10,000$ test data, while on CIRFA10 we select $1000$ examples from the dataset for evaluation.

\subsection{Evaluation of baselines} \label{evaluate_baselines}

\begin{figure}
    \begin{centering}
    \includegraphics[width=0.7\textwidth]{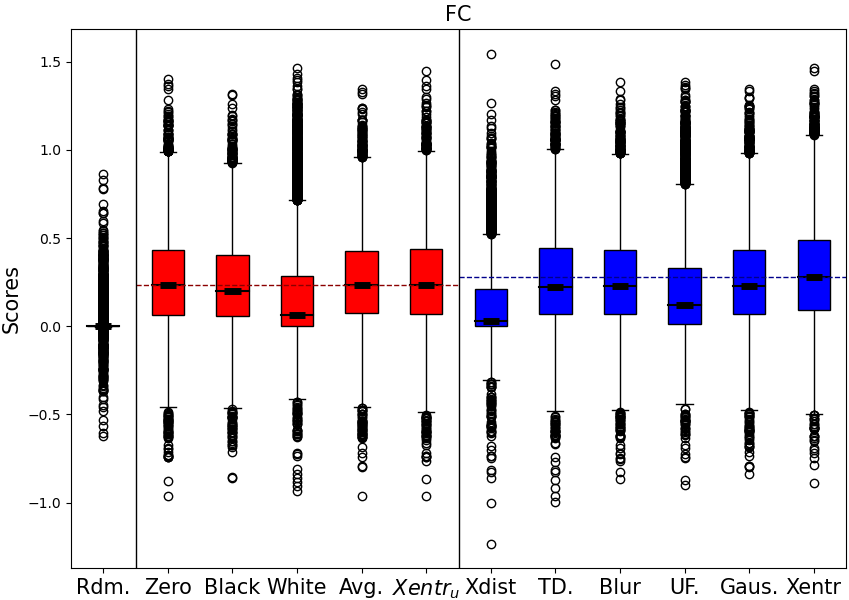}
    \caption{Ablation tests on different IG baselines of FC model. From left to right: randomly generated saliency maps (comparison reference), zero, black, white, average of current input, max entropy with uniform values, max distance, average of training data, blurred, uniform, Gaussian noise and max entropy baselines, respectively. The red boxes represent baselines with uniform values on all pixels, while the blue boxes are free of this restriction. The bolded black bar in the box is the median, and the horizontal dashed line indicates the optimal median value of the baseline in the current category (uniform or non-uniform).}
    \label{eval_fc_box}
    \end{centering}
\end{figure}

We compare the performances of all possible IG baselines by the new ablation test proposed in section \ref{Methods: Xentr ablation} and illustrated in the box plot of Figure \ref{eval_fc_box} and Table \ref{table_fc}. The model utilized in this figure is FC, in order to demonstrate the impacts of various types of models, the same experiments are conducted with other models and exhibited in Figure \ref{eval_cnn_box} and Figure \ref{eval_resnet_box}. The results reveal that $Xentr_{u}$ and $Xentr$ are equal or superior to other baselines in the same categories. Interestingly, the Max Distance baseline consistently performs worse than the remaining ones in the non-uniform baseline. There is extensive information about the input instances in the Max Distance baseline, including object boundaries, gray value information (see Figure \ref{visu_diff_baselines}), which validates our view that the more information about current instances is contained in the baseline, the less is integrated by IG and thus the generated explanation is weaker in terms of credibility.

\subsection{Evaluation of explainability methods} \label{evaluate_exp_meds}

\begin{figure*}
    \begin{centering}
    \includegraphics[width=1\textwidth]{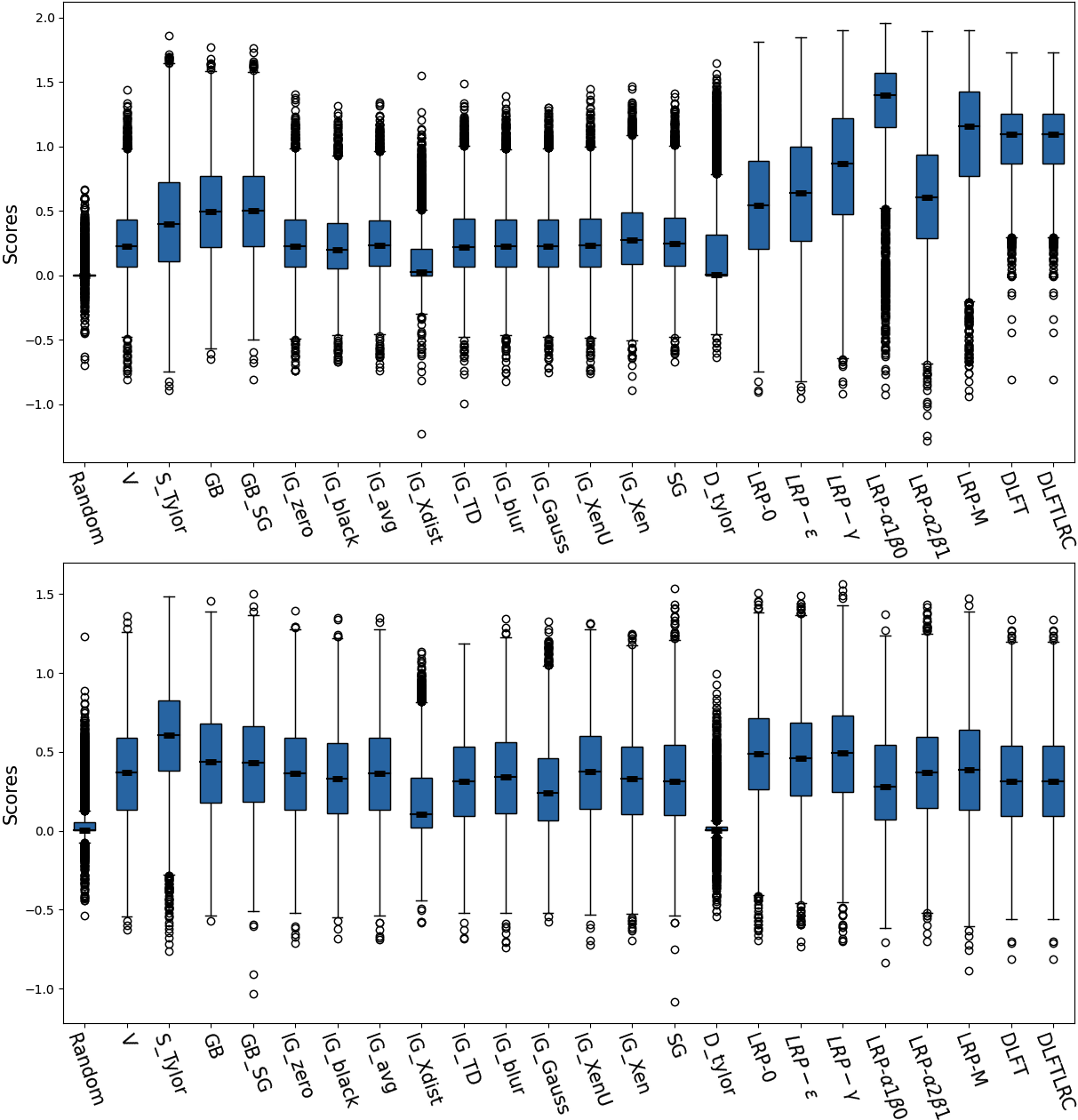}
    \caption{Ablation tests for different explainability methods. The upper and lower parts demonstrate the evaluation results on FC and CNN models, respectively. From left to right: randomly generated saliency maps (comparison reference), Vanilla Gradients, Simple Taylor Decomposition, Guided Back-propagation, Guided Back-propagation with SmoothGrad, Integrated Gradients with zero, black, average of current instance, Max distance, average of training data, blurred, Gaussian noise, (uniform) maximum entropy baselines, SmoothGrad, Deep Taylor Decomposition, LRP-$0$, LRP-$\epsilon$, LRP-$\gamma$, LRP-$\alpha\beta$, LRP-Composite, DeepLIFT and DeepLIFT with Linear Rule $\&$ Reveal Cancel, respectively. The bolded black bar in the box is the median value.}
    \label{eval_explainability_methods}
    \end{centering}
\end{figure*}

We compare multiple gradient-based explainability methods as well with our enhanced ablation tests, and report the results in Figure \ref{eval_explainability_methods}, Table \ref{all_compare_fc} and \ref{all_compare_cnn}. For IG, we excluded the white and Xdist baselines that performed unsatisfactorily in Section \ref{evaluate_baselines}. As a reference, a randomly generated saliency map is also involved in the evaluation and drawn in the first box. It can be seen that the random one fails almost to diminish the entropy of the model prediction (median $\approx 0$), whereas all other explainability methods are valid (median $>0$). Interestingly, We find that explainability approaches perform discrepantly on different models. For example, in FC, $LRP-\alpha1\beta0$ significantly outperforms other explainability methods, while in CNN, the performance is lower than the average of the LRP series.

%% file: 6_Conclusion.tex
\section{Conclusion}
This work identifies the conservation deficiencies of the existing IG baselines and ablation tests from an informational perspective, and proposes a new baseline and an enhanced ablating evaluation method based on the "missingness" necessitated by the explainability approaches. However, we acknowledge that existing ablation tests are still controversial in terms of, for example, feature correlation. In future work, we will direct our efforts to investigating more persuasive assessments for explainability methods.

%% file: 7_Supplementary.tex
\clearpage
\section{Supplementary Material}
\beginsupplement

\subsection{Hybrid explainer} \label{hybrid_explainer}

To approximate the ground truth explanations, we incorporated multiple explainability methods to vote on attributions. Our hybrid explainer contains $17$ explainability methods based on back-propagation, Taylor decomposition, LRP series and DeepLIFT series, as shown in Table \ref{hybrid-explainer}. We generate saliency maps separately for each data in the test set using all explainers, and then obtain individual scores of the explainers based on the attributions by employing a simple ablating evaluation test. Subsequently, we normalized the matrix of scores for each explainer on the test set, which is considered as the weights of the participating explainers. While comparing ground truth explanations, for each instance, we first obtained the corresponding saliency map of each explainer, and then weight those maps according to the aforementioned matrix. Such a weighted saliency map can be considered as an approximation of the ground truth explanation integrated according to the ablation scores of each explainability method.

\begin{table*}[]
\centering
\begin{tabular}{cccc}
\hline
Back-propagation-based & Taylor Decomposition & LRP-series                                    & DeepLIFT-series \\ \hline
Vanilla Gradients\cite{simonyan2014deep}      & Simple TD\cite{montavon2017explaining}            & LRP-0\cite{bach2015pixel}                                         & DeepLIFT\cite{shrikumar2017learning}        \\
GB\cite{springenberg2014striving}                     & Deep TD\cite{montavon2017explaining}              & LRP-$\epsilon$\cite{bach2015pixel}                  & DeepLIFT-LRRC\cite{shrikumar2017learning}   \\
GB+SG\cite{smilkov2017smoothgrad}                  &                      & LRP-$\gamma$\cite{bach2015pixel}                     &                 \\
SG\cite{smilkov2017smoothgrad}                     &                      & LRP-$\alpha\beta$\cite{bach2015pixel} &                 \\
IG\cite{pmlr-v70-sundararajan17a}                     &                      & LRP-Composite\cite{lapuschkin2017understanding}                                 &                 \\
IG+SG\cite{goh2021understanding}                  &                      &                                  &                 \\ \hline
\end{tabular}
\caption{Components of the hybrid explainer, where GB, SG, IG, TD and LRRC stand for Guided Back-propagation, SmoothGrad, Integrated Gradients, Taylor Decomposition and Linear Rule \& RevealCancel}\label{hybrid-explainer}
\end{table*}

\subsection{Visualization of additional inputs} \label{viasuliza_add_inputs}

\begin{figure*}
    \begin{centering}
    \includegraphics[width=1\textwidth]{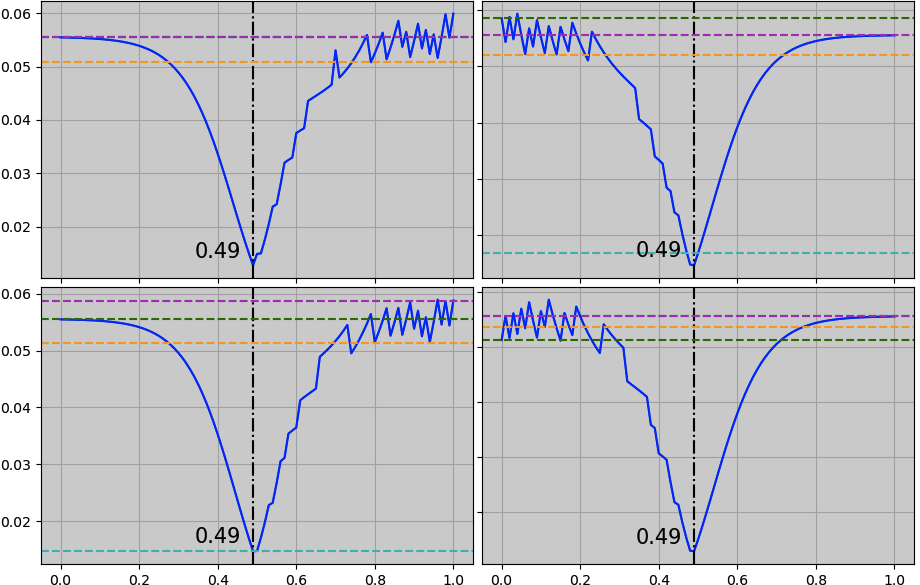}
    \caption{KL-loss curves for four different inputs. The input vectors from top left to bottom right are [0,0], [0,1], [1,0] and [1,1] respectively. We also marked the loss of zero (green), black (cyan), white (purple) and random (orange) baselines (partially obscured by each other, but significantly higher than the loss of $B_{Xentr}$).}
    \label{sup_4_diff_vectors}
    \end{centering}
\end{figure*}

We complement in figure \ref{sup_4_diff_vectors} with three additional different input vectors for Figure \ref{toydata_res}: i.e. [0,1], [1,0] and [1,1]. It can be observed that the minimum of the entropy curve is not impacted by the input vector.

\subsection{Additional evaluation results of baselines} \label{additional_eval_baseline}

\subsubsection{Figure results}
\begin{figure}
    \begin{centering}
    \includegraphics[width=0.8\textwidth]{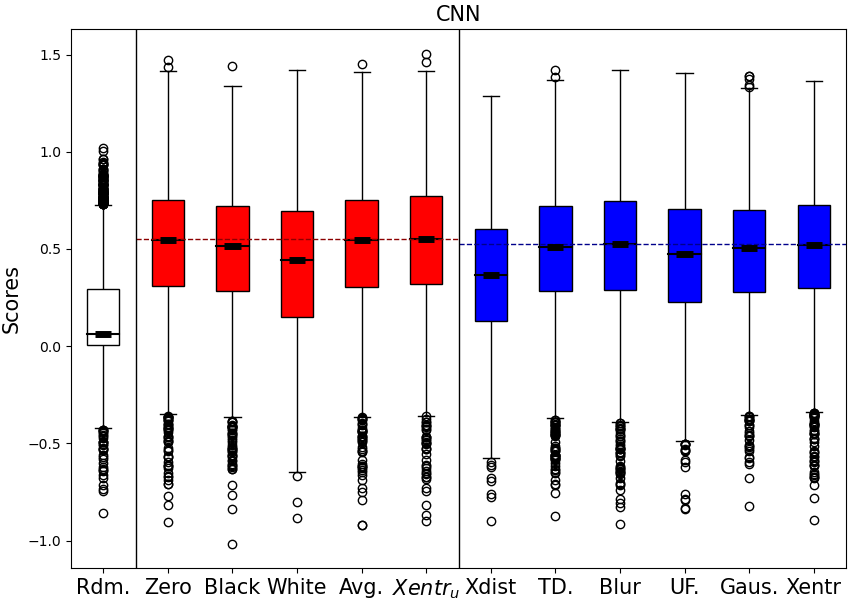}
    \caption{Ablation tests on different IG baselines of CNN model. From left to right: randomly generated saliency maps (comparison reference), zeros, black, white, average of current input, max entropy (uniform value), max distance, average of training data, blurred, uniform, Gaussian noise and max entropy baselines, respectively. The red boxes represent baselines with uniform values on all pixels, while the blue boxes are free of this restriction. The bolded black bar in the box is the median, and the horizontal dashed line indicates the optimal median value of the baseline in the current category (uniform or non-uniform).}
    \label{eval_cnn_box}
    \end{centering}
\end{figure}

\begin{figure}
    \begin{centering}
    \includegraphics[width=0.8\textwidth]{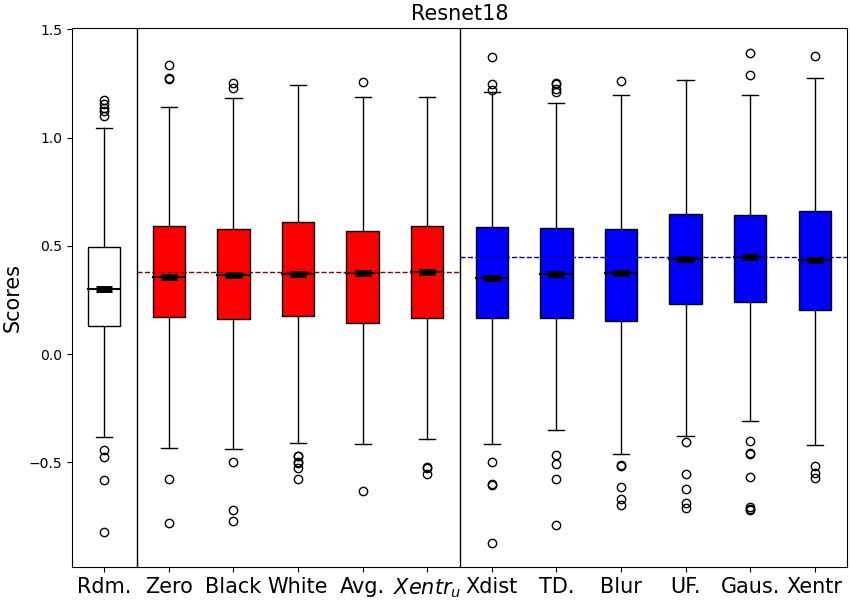}
    \caption{Ablation tests on different IG baselines of Resnet18 model.}
    \label{eval_resnet_box}
    \end{centering}
\end{figure}

Figures \ref{eval_cnn_box} and \ref{eval_resnet_box} supplement two evaluation experiments with different models, CNN and ResNet18. The proposed baselines perform beyond most baselines as well.

\subsubsection{Tabular results}

\begin{table}[]
\centering
\begin{tabular}{c|c|ccccc|cccccc}
\hline
Scores        & Random & \multicolumn{5}{c|}{Uniform Baselines}                          & \multicolumn{6}{c}{Non-uniform Baselines}          \\ \hline
              & Rdm.   & Zero  & Black & White & Avg.  & $X_{entr_u}$ & Xdist & TD.   & Blur  & UF.   & Gaus. & $X_{entr}$                      \\ \hline 
$\bar{S}$     & 0.010  & 0.272 & 0.250 & 0.188 & 0.273 & 0.277        & 0.131 & 0.274 & 0.274 & 0.210 & 0.274 & \textbf{0.321} \\
$\tilde{S}$   & 0.000  & 0.232 & 0.201 & 0.066 & 0.234 & 0.235        & 0.029 & 0.225 & 0.228 & 0.124 & 0.228 & \textbf{0.279} \\
$\sigma^2(S)$ & 0.003  & 0.066 & 0.058 & 0.077 & 0.062 & 0.067        & 0.042 & 0.067 & 0.065 & 0.068 & 0.066 & 0.077                           \\ \hline
\end{tabular}
\caption{Tabular results for comparison of different baselines on the FC model. From left to right: randomly generated saliency maps (comparison reference), zeros, black, white, average of current input, max entropy (uniform value), max distance, average of training data, blurred, uniform, Gaussian noise and max entropy baselines, respectively. Where $\bar{S}$, $\tilde{S}$ and $\sigma^2(S)$ denote the average, median and variance respectively.}\label{table_fc}
\end{table}

\begin{table}[]
\centering
\begin{tabular}{c|c|ccccc|cccccc}
\hline
Scores        & Random & \multicolumn{5}{c|}{Uniform Baselines}                          & \multicolumn{6}{c}{Non-uniform Baselines}          \\ \hline
              & Rdm.   & Zero  & Black & White & Avg.  & $X_{entr_u}$                    & Xdist & TD.   & Blur  & UF.   & Gaus. & $X_{entr}$ \\ \hline
$\bar{S}$     & 0.160  & 0.531 & 0.504 & 0.446 & 0.528 & \textbf{0.542} & 0.382 & 0.503 & 0.517 & 0.471 & 0.493 & 0.511      \\
$\tilde{S}$   & 0.060  & 0.544 & 0.513 & 0.444 & 0.546 & \textbf{0.553} & 0.366 & 0.509 & 0.524 & 0.474 & 0.503 & 0.521      \\
$\sigma^2(S)$ & 0.045  & 0.093 & 0.087 & 0.105 & 0.092 & 0.096                           & 0.083 & 0.091 & 0.096 & 0.093 & 0.084 & 0.088      \\ \hline
\end{tabular}
\caption{Tabular results for comparison of different baselines on the CNN model.}\label{table_cnn}
\end{table}

\begin{table}[]
\centering
\begin{tabular}{c|c|ccccc|cccccc}
\hline
Scores        & Random & \multicolumn{5}{c|}{Uniform Baselines}       & \multicolumn{6}{c}{Non-uniform Baselines}                                    \\ \hline
              & Rdm.   & Zero  & Black & White & Avg.  & $X_{entr_u}$ & Xdist & TD.   & Blur  & UF.   & Gaus.                           & $X_{entr}$ \\ \hline
$\bar{S}$     & 0.322  & 0.383 & 0.381 & 0.396 & 0.382 & 0.390        & 0.379 & 0.382 & 0.379 & 0.447 & \textbf{0.451} & 0.445      \\
$\tilde{S}$   & 0.300  & 0.354 & 0.365 & 0.370 & 0.372 & 0.378        & 0.353 & 0.370 & 0.373 & 0.440 & \textbf{0.446} & 0.433      \\
$\sigma^2(S)$ & 0.066  & 0.082 & 0.080 & 0.087 & 0.083 & 0.081        & 0.085 & 0.080 & 0.084 & 0.083 & 0.084                           & 0.094      \\ \hline
\end{tabular}
\caption{Tabular results for comparison of different baselines on the ResNet18 model. Although the Gaussian baseline outperforms the others in this experiment, the Maximum Entropy baselines are not significantly inferior.}\label{table_resnet18}
\end{table}

\begin{table}[]
\centering
\resizebox{.4\textwidth}{!}{
\begin{tabular}{cccc}
\hline
           & $\bar{S}$              & $\tilde{S}$              & $\sigma^2(S)$     \\ \hline
Rdm.       & 0.010          & 0.000          & 0.003 \\
V          & 0.269          & 0.222          & 0.062 \\
S\_Tylor   & 0.456          & 0.397          & 0.156 \\
GB         & 0.517          & 0.491          & 0.130 \\
GB\_SG     & 0.520          & 0.498          & 0.127 \\
IG\_zero   & 0.272          & 0.227          & 0.062 \\
IG\_black  & 0.250          & 0.197          & 0.055 \\
IG\_avg    & 0.273          & 0.230          & 0.059 \\
IG\_Xdist  & 0.130          & 0.026          & 0.039 \\
IG\_TD     & 0.274          & 0.220          & 0.063 \\
IG\_blur   & 0.274          & 0.223          & 0.062 \\
IG\_Gaus   & 0.275          & 0.222          & 0.063 \\
IG\_Xentru & 0.277          & 0.230          & 0.063 \\
IG\_Xentr  & 0.319          & 0.275          & 0.075 \\
SG         & 0.288          & 0.243          & 0.063 \\
D\_Tylor   & 0.203          & 0.003          & 0.110 \\
LRP-0      & 0.573          & 0.540          & 0.188 \\
LRP-$\epsilon$      & 0.648          & 0.635          & 0.212 \\
LRP-$\gamma$      & 0.829          & 0.868          & 0.233 \\
LRP-$\alpha1\beta0$    & \textbf{1.283} & \textbf{1.396} & 0.187 \\
LRP-$\alpha2\beta1$    & 0.618          & 0.605          & 0.186 \\
LRP-M      & 1.054          & 1.156          & 0.229 \\
DLFT       & 1.032          & 1.096          & 0.096 \\
DLFTLR     & 1.032          & 1.096          & 0.096 \\ \hline
\end{tabular}}
\caption{Tabular results for comparison of different explainability methods on the FC model. From top to bottom: randomly generated saliency maps (comparison reference), Vanilla Gradients, Simple Taylor Decomposition, Guided Back-propagation, Guided Back-propagation with SmoothGrad, Integrated Gradients with zero, black, average of current instance, Max distance, average of training data, blurred, Gaussian noise, (uniform) maximum entropy baselines, SmoothGrad, Deep Taylor Decomposition, LRP-$0$, LRP-$\epsilon$, LRP-$\gamma$, LRP-$\alpha\beta$, LRP-Composite, DeepLIFT and DeepLIFT with Linear Rule $\&$ Reveal Cancel, respectively. Where $\bar{S}$, $\tilde{S}$ and $\sigma^2(S)$ denote the average, median and variance respectively.}\label{all_compare_fc}
\end{table}

\begin{table}[]
\centering
\resizebox{.4\textwidth}{!}{
\begin{tabular}{cccc}
\hline
           & $\bar{S}$              & $\tilde{S}$              & $\sigma^2(S)$     \\ \hline
Rdm.       & 0.054          & 0.005          & 0.014 \\
V          & 0.381          & 0.369          & 0.078 \\
S\_Tylor   & \textbf{0.598} & \textbf{0.605} & 0.096 \\
GB         & 0.449          & 0.435          & 0.098 \\
GB\_SG     & 0.438          & 0.431          & 0.088 \\
IG\_zero   & 0.379          & 0.364          & 0.080 \\
IG\_black  & 0.354          & 0.331          & 0.075 \\
IG\_avg    & 0.379          & 0.363          & 0.079 \\
IG\_Xdist  & 0.200          & 0.105          & 0.053 \\
IG\_TD     & 0.338          & 0.314          & 0.074 \\
IG\_blur   & 0.358          & 0.343          & 0.077 \\
IG\_Gaus   & 0.286          & 0.241          & 0.063 \\
IG\_Xentru & 0.386          & 0.373          & 0.081 \\
IG\_Xentr  & 0.342          & 0.326          & 0.070 \\
SG         & 0.348          & 0.310          & 0.081 \\
D\_Tylor   & 0.033          & 0.002          & 0.009 \\
LRP-0      & 0.493          & 0.488          & 0.098 \\
LRP-$\epsilon$      & 0.466          & 0.457          & 0.098 \\
LRP-$\gamma$      & 0.498          & 0.491          & 0.106 \\
LRP-$\alpha1\beta0$    & 0.330          & 0.279          & 0.081 \\
LRP-$\alpha2\beta1$    & 0.391          & 0.368          & 0.084 \\
LRP-M      & 0.408          & 0.383          & 0.097 \\
DLFT       & 0.338          & 0.312          & 0.070 \\
DLFTLR     & 0.338          & 0.312          & 0.070 \\ \hline
\end{tabular}}
\caption{Tabular results for comparison of different explainability methods on the CNN model.}\label{all_compare_cnn}
\end{table}